\documentclass{article}
\usepackage{amssymb}
\usepackage{graphicx}
\usepackage{subcaption}
\usepackage{arydshln}
\usepackage{xcolor}
\usepackage{booktabs}
\usepackage{amsmath}
\usepackage[table,xcdraw]{xcolor}
\usepackage{mathrsfs}
\usepackage{wrapfig}
\usepackage{caption}
\usepackage{float}
\usepackage{placeins}
\usepackage{enumitem}
\usepackage{multirow}    
\usepackage{hyperref}
\usepackage{appendix}

\usepackage[final]{corl_2025} 

\title{TA-VLA: Elucidating the Design Space of Torque-aware Vision-Language-Action Models}

\author{
Zongzheng Zhang\footnotemark[1] $^{\ 1}$, Haobo Xu\footnotemark[1] $^{\ 2}$, Zhuo Yang\footnotemark[1] $^{\ 1}$, \\
\textbf{Chenghao Yue$^{1}$, Zehao Lin$^1$,}
\textbf{Huan-ang Gao$^1$, Ziwei Wang$^3$, Hao Zhao\footnotemark[2] $^{\ 1,2}$}\vspace{0.2cm}\\
\footnotemark[1]\ \ Equal Contribution;  \footnotemark[2]\ \ Corresponding author \vspace{0.2cm}\\
    $^1$ 
  Beijing Academy of Artificial Intelligence, BAAI\\
    $^2$ Institute for AI Industry Research (AIR), Tsinghua Univeristy \\
    $^3$ Nanyang Technological University \vspace{0.15cm}\\
  \texttt{zhaohao@air.tsinghua.edu.cn} \vspace{0.15cm}\\
  \href{https://zzongzheng0918.github.io/Torque-Aware-VLA.github.io/}{https://zzongzheng0918.github.io/Torque-Aware-VLA.github.io/}\vspace{-0.3cm}
}

\begin{document}

\maketitle

\vspace{-20pt}
\begin{abstract}
    Many robotic manipulation tasks require sensing and responding to force signals such as torque to assess whether the task has been successfully completed and to enable closed-loop control. However, current Vision-Language-Action (VLA) models lack the ability to integrate such subtle physical feedback. In this work, we explore Torque-aware VLA models, aiming to bridge this gap by systematically studying the design space for incorporating torque signals into existing VLA architectures. We identify and evaluate several strategies, leading to three key findings. First, introducing torque adapters into the decoder consistently outperforms inserting them into the encoder.
    This is because torque signals align more closely with the decoder’s input, and the decoder is more sensitive to variations in input.
    Second, torque history proves to be a critical signal. We find that the most effective way to incorporate it is by summarizing the entire history into a single token, as this preserves the original input pattern of the decoder.
    Third, inspired by joint prediction and planning paradigms in autonomous driving, we propose predicting torque as an auxiliary output, which further improves performance. This strategy encourages the model to build a physically grounded internal representation of interaction dynamics. Extensive quantitative and qualitative experiments across contact-rich manipulation benchmarks validate our findings. 
\end{abstract}

\vspace{-5pt}

\section{Introduction}

\vspace{-5pt}
\begin{figure}
    \centering
    \includegraphics[width=\linewidth]{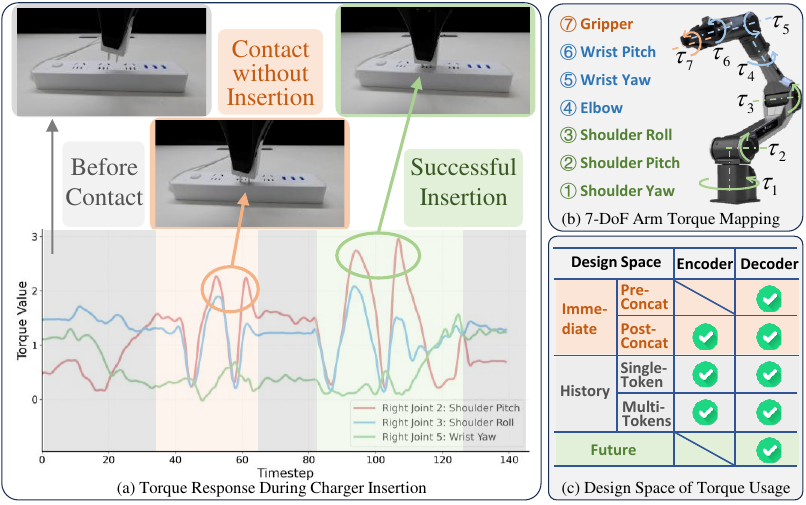}
    \caption{(a) Torque response of the 7-DoF arm during a charger-insertion task. Shaded \textcolor{gray}{gray} regions mark periods of no contact, where torques remain nearly flat. The \textcolor{orange}{orange}-tinted segment shows a failed insertion attempt—contact is made but the plug does not enter the socket, producing only small torque fluctuations. The \textcolor[RGB]{100,180,100}{green}-tinted segment highlights a successful insertion, characterized by large, distinctive torque spikes as the plug seats fully.
(b) Visualization of the 7-DoF robot arm, highlighting joint torque mappings. 
(c) Design space of torque-based features explored in this work, spanning current, historical, and future signals.}
    \vspace{-0.5cm}
    \label{teaser}
\end{figure}

Understanding physical interactions through force cues is essential for mastering real-world robotic manipulation. One particularly informative signal is joint torque, which reflects subtle variations in end-effector contact dynamics without requiring external force sensors~\citep{likar2014external, shan2023fine, xu2024end}. As shown in Figure~\ref{teaser}(a), different outcomes in a seemingly simple task like charger insertion—no contact, failed insertion, and successful plug-in—can be clearly distinguished by the joint torque profiles of a 7-DoF arm. These torque responses offer rich physical context that is otherwise imperceptible from RGB observations alone. However, despite the growing success of Vision-Language-Action (VLA) models ~\citep{kim2024openvla, octo, pi0, RDT, pi0.5} in bridging vision and control, their ability to interpret and leverage such physical feedback remains limited. Our work aims to bridge this gap by integrating torque signals into pretrained VLA models, enabling contact-sensitive decision-making without compromising generalization or scalability.

The challenge lies in how to embed torque into VLA architectures. Torque is a proprioceptive signal, structurally different from image and language inputs, and varies across time, especially during contact-rich phases. As illustrated in Figure~\ref{teaser}(c), multiple torque integration strategies exist across three axes—when (immediate vs. historical vs. predictive), where (encoder vs. decoder), and how (single token vs. multi-token). These options form a broad design space, but lack systematic understanding. Our motivation is thus twofold: (i) identify the most effective design choices for torque-aware VLA models, and (ii) derive generalizable principles that guide future integration of physical modalities.

\textbf{Our first insight is that torque signals should be integrated into the decoder, not the encoder.} As validated through HSIC analysis ~\citep{hsic} and ablation (Sec.~\ref{subsec:obs_where}), decoder-side integration aligns torque with other proprioceptive signals (like joint angles) during action generation. This placement leverages the decoder’s higher sensitivity to fine-grained variations—critical in contact-rich scenarios (e.g., distinguishing a failed vs. successful plug in Figure~\ref{teaser}(a)).

\textbf{Our second insight is that historical torque information is more informative than single-frame input.} However, injecting multiple tokens can disrupt the decoder’s learned input patterns. We find that encoding the entire torque history into a single token in the decoder (Figure~\ref{fig:his}(c)) balances informativeness with architectural stability. This design choice outperforms both per-frame and encoder-side history integration (Sec.~\ref{subsec:obs_his}), enabling robust temporal modeling of contact dynamics, as seen during insertion and retries in Figure~\ref{fig:visualization}.

\textbf{Our third insight is that predicting future torque alongside actions helps build a physically grounded latent space.} Inspired by multi-task architectures in autonomous driving~\citep{MTR}, we propose a unified action–torque diffusion model (Sec.~\ref{sec:obj}), which allows the policy not only to act but also to anticipate physical consequences (see prediction curves in Figure~\ref{fig:eff_pred}). This auxiliary task encourages the model to internalize contact dynamics beyond observation alone.

Finally, we validate our full system with extensive real-world experiments across 10 diverse tasks—including five contact-rich ones where torque feedback is critical. Our final model ($\pi_0$+obs+obj) achieves consistent gains over strong VLA baselines~\citep{ACT, RDT, pi0} (Table~\ref{tab:Quantitative Results}), and generalizes across both model architectures and robot embodiments. These results confirm that torque-aware VLA models not only improve task success but also increase robustness and generalization.

In summary, our contributions are: (1) We propose a systematic design space for torque-aware VLA modeling, spanning where and how torque is integrated (Figure~\ref{teaser}(c)). (2) We find that decoder-side, single-token torque history yields the best proprioceptive alignment and performance. (3) We introduce a unified action–torque diffusion model that enables anticipatory learning through torque prediction. (4) We demonstrate significant performance gains on contact-rich manipulation tasks across models and embodiments.


\vspace{-9pt}
\section{Related Work}
\vspace{-6pt}
\textbf{Vision-Language-Action Model.} Recently, Large Language Models (LLMs) \citep{gpt4,deepseekllm,llama3,qwen2.5} and Vision-Language Models (VLMs) \citep{gpt4o,llava1.5,llavaonevision,qwen2.5vl,liu2023delving,li2022toist,ding2024hint,jin2024tod3cap} have achieved remarkable success, while generative models have enabled continuous outputs such as image generation \citep{diffusion,vqvae,flowmatching, zhang2025chameleon}. These techniques pave the way for the advent of VLA models, which combine visual perception, language understanding, and action generation abilities, and demonstrate strong generalizability, utilizing millions of training data samples including different tasks and devices \citep{Cogact2024,huang2023embodied,li2023vision,spatialvla2025,tinyvla, jiang2025diffvla,chi2025impromptuvla}. Recent works regarding VLA models can be categorized into several modes based on action generation methods, including diffusion policy-based models \citep{octo,RDT}, flow matching-based models \citep{pi0}, and autoregressive generation models \citep{RT-2,kim2024openvla,pertsch2025fast}. For example, Octo \citep{octo} and RDT-1B \citep{RDT} utilize a diffusion head and a transformer backbone to predict actions, while $\pi_0$ \citep{pi0} uses conditional flow matching to generate high-frequency action sequences. Models with hybrid architectures further combine multiple generative techniques to leverage the strengths of each. For instance, HybridVLA \citep{hybridvla} merges diffusion and autoregressive approaches within a single model.

\textbf{Imitation Learning with Force/Torque.} While most of the existing work regarding imitation learning utilizes joint-position and visual information \citep{pi0,pi0.5}, force/torque information as an extra input is gaining more and more attention. Recent studies have demonstrated that force/torque signals could equip controlling policies with the ability to handle a wide range of real-world tasks, which include subtle and precise manipulation \citep{chen2022visuo,liu2024forcemimic,ding2024preafford,hou2024adaptive,chen2025dexforce, van2024built}. From the perspective of sources of force/torque, most approaches rely on additional sensors to obtain 6D wrench measurements \citep{he2024foar,aburub2024learning,huang20243d,kamijo2024learning,van2024built}, which leads to higher economic costs and limitations in harsh operating conditions. 
Furthermore, although some works are trying to incorporate force/torque information with visual and text inputs, they typically train policies from scratch and fail to leverage the advantages of pretrained VLA models \citep{xue2025reactive,wu2024tacdiffusion,kobayashi2025bi,li2024haptic}. For example, FACTR \citep{liu2025factr} requires a complex training pipeline to align different modalities and lacks flexibility. In contrast, incorporating the force modality into pretrained VLA models offers two major benefits. 1) VLA models already possess a strong foundation in cross-modal learning, having been trained on large-scale datasets; thus, integrating an additional modality is easier. 2) VLA models typically learn shared feature representations across modalities, making it more efficient to accommodate new modality. In this work, we systematically explore ways to enrich pretrained VLA models with force information, allowing them to act as world models that accurately perceive and predict the environment through a unified understanding of vision, force, and instruction over historic, immediate, and future states.





\vspace{-9pt}
\section{Torques are Good Indicators for End-effector Status}
\vspace{-6pt}
\label{sec:method}

In robotic manipulation, external contact at the end-effector induces mechanical responses across the entire kinematic chain. These responses manifest as observable variations in joint torques. In this section, we formalize how joint torque signals encode contact force information via the mechanical arm's differential kinematics and dynamics.

\textbf{Formulation.} Suppose that the manipulator has $n $ degrees of freedom, with joint configuration vector $\boldsymbol{q} \in \mathbb{R}^n$. The full-body dynamics in the presence of external contact are given by:
\begin{equation}
\label{eq:dynamics}
\boldsymbol{M}(\boldsymbol{q}) \ddot{\boldsymbol{q}} + \boldsymbol{C}(\boldsymbol{q}, \dot{\boldsymbol{q}})\dot{\boldsymbol{q}} + \boldsymbol{G}(\boldsymbol{q}) = \boldsymbol{\tau}_{\text{cmd}} + \boldsymbol{\tau}_{\text{ext}},
\end{equation}
where $\boldsymbol{\tau}_{\text{cmd}} \in \mathbb{R}^n$ is the commanded torque, and $\boldsymbol{\tau}_{\text{ext}} \in \mathbb{R}^n$ is the torque contribution caused by external forces applied at the end-effector. The term \( \boldsymbol{M}(\boldsymbol{q}) \in \mathbb{R}^{n \times n} \) is the inertia matrix, \( \boldsymbol{C}(\boldsymbol{q}, \dot{\boldsymbol{q}}) \in \mathbb{R}^{n \times n} \) is the Coriolis and centrifugal force matrix, and \( \boldsymbol{G}(\boldsymbol{q}) \in \mathbb{R}^n \) is the gravity torque vector. Here, \( \dot{\boldsymbol{q}} \in \mathbb{R}^n \) is the joint angular velocity vector, and \( \ddot{\boldsymbol{q}} \in \mathbb{R}^n \) is the joint angular acceleration vector.

\textbf{Mapping.} Assume that the end-effector makes contact with a rigid environment and experiences a spatial force (wrench) denoted as \(\boldsymbol{F}_{\text{ext}} \in \mathbb{R}^{6} \). Given that the virtual end-effector displacement is related to joint displacement by $\delta \boldsymbol{x} = \boldsymbol{J}(\boldsymbol{q}) \delta \boldsymbol{q}$, we obtain:
\begin{equation}
\boldsymbol{\tau}_{\text{ext}} = \boldsymbol{J}^\top(\boldsymbol{q}) \boldsymbol{F}_{\text{ext}} \in \mathbb{R}^n.
\label{eq:tau_ext}
\end{equation}

Here, \( \boldsymbol{J}(\boldsymbol{q}) \in \mathbb{R}^{6 \times n} \) is the Jacobian matrix, which maps the velocity from the end-effector space to the joint space. This equation is fundamental: it states that any external force acting on the end-effector is projected back into joint space through the transpose of the Jacobian matrix. Therefore, when a contact event occurs (e.g., the robot touches a surface), the resulting torque signal can be decomposed as:
\begin{equation}
\boldsymbol{\tau}_{\text{measured}} = \boldsymbol{\tau}_{\text{model}} + \boldsymbol{J}^\top(\boldsymbol{q}) \boldsymbol{F}_{\text{ext}}.
\end{equation}
Here, $\boldsymbol{\tau}_{\text{measured}}$ is the observed joint torque, and $\boldsymbol{\tau}_{\text{model}}$ accounts for the expected torque due to internal dynamics. This expression shows that observing variations in joint torques allows us to infer the net external wrench acting on the end-effector, provided that the manipulator dynamics are accurately modeled.


\textbf{Conclusion.} The joint torque vector $\boldsymbol{\tau}_{\text{measured}}$ inherently carries information about external contacts through the relationship Eq.~\eqref{eq:tau_ext}. This result forms the theoretical basis for torque-based contact estimation, where joint torque deviations from the nominal model are used to infer spatial interaction forces, enabling sensorless force estimation, collision detection, and compliant manipulation.

\vspace{-9pt}
\section{Sense What Was: Torques as Observations}

\vspace{-6pt}

\label{sec:obs}
In this section, we integrate torque signals as an additional observation into the VLA framework and investigate its effects. Most VLA models consist of two main components: a \textit{conditioning encoder} and a \textit{denoising decoder}, which we refer to simply as the \textit{encoder} and the \textit{decoder}. The encoder percepts the environment, while the decoder yields actions. To be specific, the encoder processes images inputs $I$ and language instruction $L$ into a unified latent space to construct contextual representations, and 
the decoder then progressively refines noisy inputs \(\boldsymbol{\hat{A}}_{t:t+H}\) to generate action sequences $\boldsymbol{A}_{t:t+H}$. For example, RDT~\citep{RDT} applies cross-attention over visual and language features to form the conditioning and employs a denoising backbone operating on low-dimensional proprioceptive inputs and noised action chunks. Similarly, $\pi_0$~\citep{pi0} leverages a PaliGemma~\citep{beyer2024paligemma} backbone to fuse vision and language inputs, followed by an action decoder.

Using $\pi_0$ as a representative case, we explore the design space of torques as inputs with following questions: (1) where to incorporate torque signals—into conditioning encoder or denoising decoder (Sec.~\ref{subsec:obs_where}) and (2) how historical torque signals can be leveraged (Sec.~\ref{subsec:obs_his}).

\vspace{-5pt}
\subsection{Where to Embed? Conditioning Encoder vs. Denoising Decoder}
\vspace{-5pt}
\label{subsec:obs_where}
\begin{figure}

    \centering
    \includegraphics[width=1\linewidth]{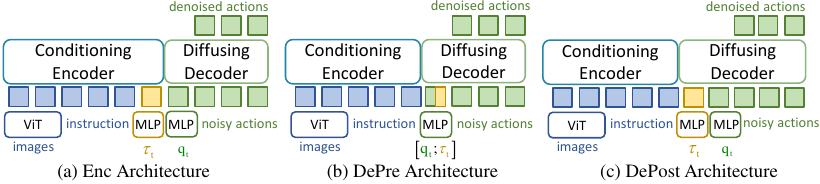}
    \vspace{-0.5cm}
    \caption{Architectures for embedding torque signals.
    }
    \label{fig:where_embed}
    \vspace{-0.5cm}
\end{figure}

At each time step \(t\), the policy \(\pi_0\) observes multiple RGB images, a textual instruction, and the robot’s joint angle state, denoted as
\(
\boldsymbol{o}_t = \bigl[\boldsymbol{I}_{t_1}, \dots, \boldsymbol{I}_{t_n},\,\boldsymbol{L}_t,\,\boldsymbol{q}_t\bigr],
\)
where \(\boldsymbol{I}_{t_i}\) is the feature vector of the \(i\)-th image, \(\boldsymbol{L}_t\) is the sequence of language tokens, and \(\boldsymbol{q}_t\) is the current robot state vector. In the original \(\pi_0\) architecture, the image features \(\{\boldsymbol{I}_{t_i}\}\) together with \(\boldsymbol{L}_t\) form the conditioning context, while \(\boldsymbol{q}_t\) is provided as a token to the denoising module.

Regarding using torque signals and integrating it into the VLA architecture, we explore two integration ways: integrating \(\tau_t\) into the encoder's inputs to leverage its multi-modality capabilities, or incorporating \(\tau_t\) into the decoder alongside \(\boldsymbol{q}_t\) to enrich the state representation. Specifically, we evaluate three possible strategies for embedding \(\tau_t\) (see Figure~\ref{fig:where_embed}):
\vspace{-0.2cm}
\begin{itemize}[leftmargin=*]
    \setlength\itemsep{0cm}{
    \item \textbf{Encoder Embedding (Enc):} encode \(\tau_t\) via an adapter into a token which is concatenated with \(\{\boldsymbol{I}_{t_1},\dots,\boldsymbol{I}_{t_n},\boldsymbol{L}_t\}\) as an extra conditioning input (Figure~\ref{fig:where_embed}(a)); 
    \item \textbf{Decoder Pre-Concatenation Embedding (DePre):} directly integrate \(\tau_t\) into the zero-padded dimensions of \(\boldsymbol{q}_t\), concatenating them to form a single combined token (Figure~\ref{fig:where_embed}(b)); 
    \item \textbf{Decoder Post-Concatenation Embedding (DePost):} encode \(\tau_t\) through an adapter and prepend the resulting token to the action expert’s state inputs (Figure~\ref{fig:where_embed}(c)).
    }
\end{itemize}
\vspace{-0.2cm}
Specifically, we employ an MLP as the torque adapter. We conducted real-world experiments on two contact-rich tasks using the three different architectures. The results are shown in Table~\ref{tab:where_to_embed}, which shows that embedding torque signals into decoder outperforms into encoder, and embedding it to a single token outperforms integrating it to the original proprioceptive state token. The reasons for the result can be summarized as follows.

\begin{wrapfigure}{r}{0.4\textwidth}
\centering
\vspace{-0.5cm}
\includegraphics[scale=0.143]{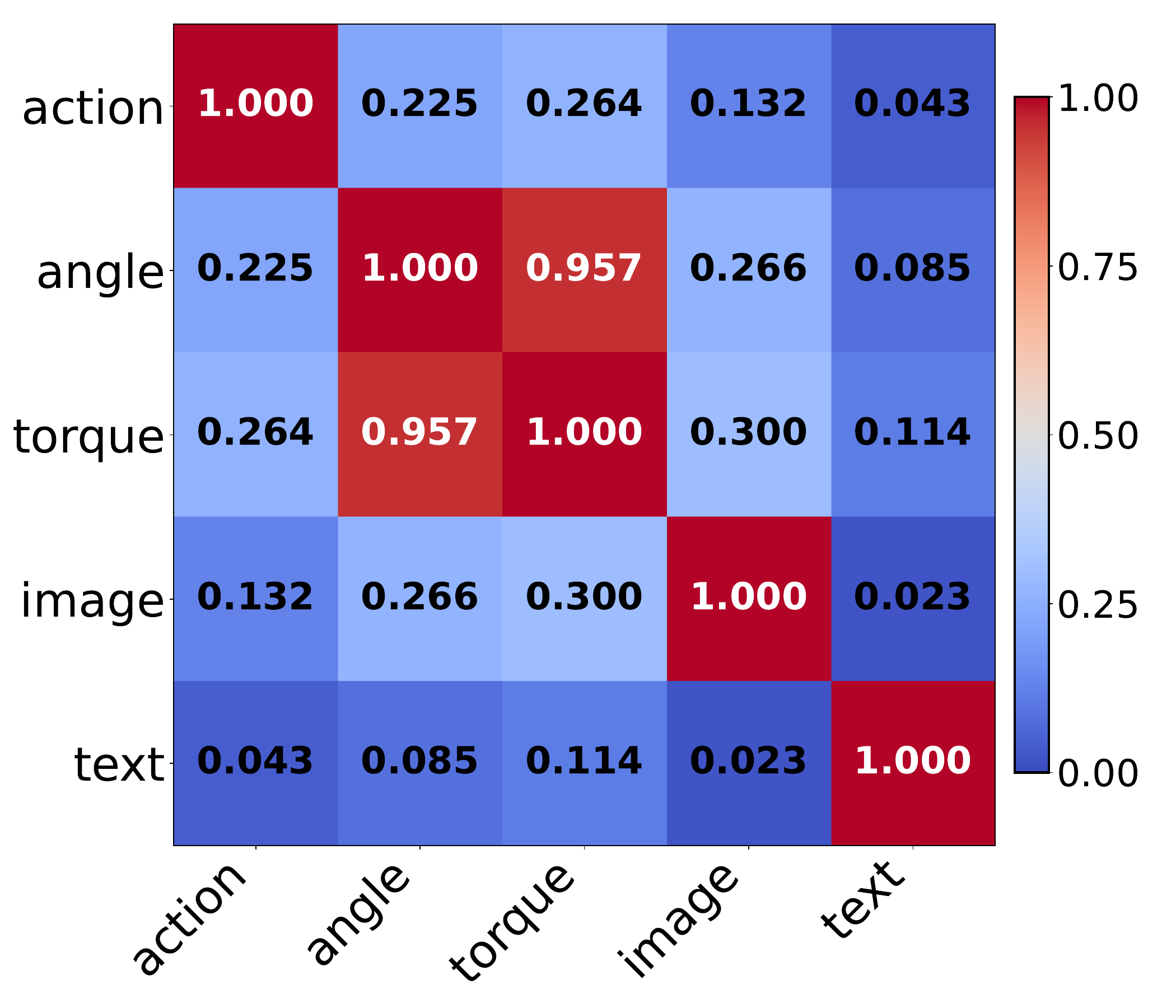}
\vspace{-6pt}
\caption{Normalized HSIC values across hidden states from different modality input tokens.}
\label{fig:hsic}
\vspace{-0.5cm}
\end{wrapfigure}

\textbf{Better Input Alignment.} Integrating torque signals $\tau_t$ into decoder outperforms placing it in encoder. Since $\tau_t$ and joint angles $q_t$ are both proprioceptive signals, fusing them during denoising better exploits their correlation, such as consistency and redundancy in contact-rich interactions. To verify this, we conduct experiments to evaluate the Normalized Hilbert-Schmidt Independence Criterion (HSIC)~\citep{hsic} values between the high-dimensional features of the inputs (together with the action) to evaluate their similarities. Figure~\ref{fig:hsic} shows that torque information is significantly more aligned with joint angle signals. Therefore, torque signals should be integrated in decoder to better enhance the proprioceptive perception.

\textbf{Sensitivity of Decoder.} The encoder, designed for diverse, ambiguous vision-language inputs, processes coarser features, whereas decoder is designed to capture subtle variations in inputs. 
To verify this, we add random noise to each input token of encoder and decoder, respectively, and evaluate the performance. Table~\ref{tab:noised} shows that, with the effect of noise, decoder shows worse performances, indicating that decoder is more sensitive to the variations in inputs; therefore, introducing $\tau_t$ to denoising enables finer utilization of subtle torque variations. Moreover, the Pre-Concatenation method significantly alters the original input token, acting as additional noise, leading to worse performance compared to the Post-Concatenation approach.
  

\begin{table}[ht]
\vspace{-0.15cm}
\centering
\begin{minipage}[t]{0.49\textwidth}
\centering
  \setlength{\tabcolsep}{3pt}
    \fontsize{7pt}{7.5pt}\selectfont
    \renewcommand{\arraystretch}{0.7}
  \begin{tabular}{l|cccc}
    \toprule
    Task & $\pi_0$& Enc&DePre&DePost\\
    \midrule
    Button Pushing & 5/20 & 7/20 & 8/20 & \textbf{10/20}\\
    \midrule
    Charger Plugging& 0/20 & 8/20 &  11/20& \textbf{12/20}\\
    \bottomrule
  \end{tabular}
  \vspace{0.1cm}
  \captionof{table}{Results of different architectures for embedding torque signals.}
  \vspace{-0.5cm}
  \label{tab:where_to_embed}
\end{minipage}
\begin{minipage}[t]{0.49\textwidth}
  \centering
  \setlength{\tabcolsep}{2pt}
    \fontsize{7pt}{7.5pt}\selectfont
    {
    \renewcommand{\arraystretch}{0.7}
  \begin{tabular}{l|ccc}
  
    \toprule
    Task & $\pi_0$ & Enc-Noised & Dec-Noised \\
    \midrule
    Bottle Pick and Place & 14/20 & 12/20 & 8/20  \\
    \midrule
    Button Pushing & 5/20 & 4/20 & 0/20  \\
    \bottomrule
  \end{tabular}
  }
  \vspace{0.1cm}
  
  \caption{Results of noised encoder and decoder with random noise.}
  \vspace{-0.8cm}

  \label{tab:noised}
\end{minipage}
\hfill
\vspace{-6pt}
\end{table}

\vspace{-3pt}
\subsection{Torque Histories Beat Single Frames}
\vspace{-3pt}
\label{subsec:obs_his}
\begin{figure}\includegraphics[scale=1]{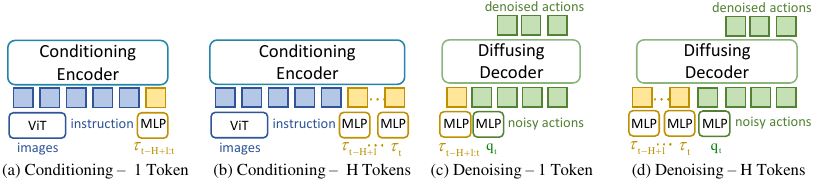}
    \caption{Architectures for embedding torque history.}
    \vspace{-0.1cm}
    \label{fig:his}
    
\end{figure}

Unlike the fixed language instruction and relatively stable visual observations---which exhibit minimal changes after end-effector contact due to occlusions---torque signals vary significantly upon contact, as illustrated in Figure~\ref{teaser}. 
To capture the dynamic patterns of torques, a single-frame torque input is insufficient. 
Encoding the history of torque signals provides the VLA model with richer patterns of physical interaction, thereby enabling better performance on contact-rich tasks.
\begin{table}[ht]
\vspace{-10pt}
\centering
\begin{minipage}[t]{0.49\textwidth}
  \centering
  
  \setlength{\tabcolsep}{2pt}
    \fontsize{7pt}{7.5pt}\selectfont
    \renewcommand{\arraystretch}{0.7}
  \begin{tabular}{l|ccccc}
    \toprule
    Task & $\pi_0$& Enc-1&Enc-H&Dec-1&Dec-H\\
    \midrule
    Button Pushing & 5/20 & 1/20 & 4/20 & \textbf{15/20}&9/20\\
    \midrule
    Charger Plugging& 0/20 & 3/20 &6/20& \textbf{16/20}&7/20\\
    \bottomrule
  \end{tabular}
  \vspace{0.1cm}
  \captionof{table}{Results of different architectures for embedding torque history.}
  \label{tab:his}
\end{minipage}
\hfill
\begin{minipage}[t]{0.49\textwidth}
  \centering
  \setlength{\tabcolsep}{2pt}
    \fontsize{7pt}{7.5pt}\selectfont
    \renewcommand{\arraystretch}{0.7}
  \begin{tabular}{l|ccc}
    \toprule
    Task & $\pi_0$ & Enc-Disrupted & Dec-Disrupted \\
    \midrule
    Bottle Pick and Place & 14/20 & 13/20 & 8/20  \\
    \midrule
    Button Pushing & 5/20 & 5/20 & 2/20  \\
    \bottomrule
  \end{tabular}
  
  \vspace{0.1cm}
  \caption{Results of disrupted encoder and decoder with extra noised tokens.}

  \label{tab:disrupted}
\end{minipage}
\hfill
   \vspace{-0.8cm}
\end{table}

To investigate the optimal way of encoding torque history, we explore two strategies: (1) frame-wise tokenization, encoding each torque frame \{$\boldsymbol{\tau}_{t-H+1}, \cdots, \boldsymbol{\tau}_{t}$\} as a  separate token, and (2) aggregate tokenization, encoding the entire history $\boldsymbol{\tau}_{t-H+1:t}$ into a single token. For completeness, we also examine whether historical torque signals should be inserted into the encoder (Figure~\ref{fig:his}(a)-(b)) or the decoder (Figure~\ref{fig:his}(c)-(d)). 
The results are shown in Table~\ref{tab:his}, indicating that encoding the entire torque history as one single token into the decoder is the best choice. The reason is as follows.


\textbf{Input Pattern Completeness.} Aggregating tokenization outperforms frame-wise tokenization, as the large number of history tokens disrupts the decoder's original input pattern completeness. To verify this hypothesis, we add extra noise tokens to the encoder and decoder, respectively. As shown in Table~\ref{tab:disrupted}, the added noise tokens easily disrupt the perception ability of the decoder. This also holds when adding extra torque history tokens, which may interfere with the patterns the decoder learned during pretraining. Therefore, even if fewer history tokens may lead to information loss, the effect of disrupting the decoder's state patterns dominates the trade-off. Moreover, in Table~\ref{tab:disrupted}, the encoder shows robustness to changes in input patterns and performs better with multiple history tokens which contain more information. However, as mentioned in Sec.~\ref{subsec:obs_where}, encoding torque signals provides advantages in proprioceptive alignment and finer-grained perception; therefore, a single token of history information to the decoder outperforms other methods.


%



\vspace{-9pt}
\section{Predict What will Be: Torques as Objectives}
\label{sec:obj}
\vspace{-6pt}

\begin{wrapfigure}{r}{0.27\textwidth}
\vspace{-1.7cm}
    \centering
    \includegraphics[scale=1]{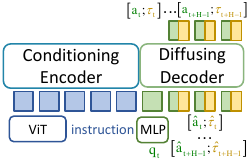}
    \vspace{-15pt}
    \caption{Architectures for Action-Torque Diffusion.}
    \label{fig:future curve}
    \vspace{-0.4cm}
\end{wrapfigure}

\textbf{Motivation.} Current VLA policies treat modalities purely as observations, missing the opportunity to internalise the robot’s own interaction dynamics. Inspired by multi-task planning in autonomous driving~\citep{MTR} and by our findings that torque information is a strong proprioceptive cue in Sec.~\ref{sec:obs}, we propose to \textbf{predict future torques together with future actions}. This auxiliary task nudges the model to build a physically grounded latent space, leading to more reliable contact-rich manipulation. 

\textbf{Unified Loss for Action–Torque Diffusion.} We now describe in detail how we train the model to predict both actions and torques, maintaining individual losses but sharing diffusion weights for efficiency.
Let \(
\boldsymbol{A}_t \in\mathbb{R}^{H\times d_a}\) denote the action chunk \( [a_t,\dots,a_{t+H-1}] \) and \(\boldsymbol{T}_t \in\mathbb{R}^{H\times d_\tau}\) denote the torque chunk 
\( [\tau_t,\dots,\tau_{t+H-1}]
\). 
The clean joint token can be expressed by \(
\boldsymbol{Z}_t = [\boldsymbol{A}_t;\,\boldsymbol{T}_t]\in\mathbb{R}^{H\times (d_a + d_\tau)}
\).
We sample Gaussian noise \(\boldsymbol{\epsilon}\sim\mathcal{N}(\boldsymbol{0},\boldsymbol{I})\) and a time step \(\alpha\in[0,1]\), forming a noisy input
\(\boldsymbol{Z}_t^\alpha = \alpha\,\boldsymbol{Z}_t + (1-\alpha)\,\boldsymbol{\epsilon}
\). To ensure both action and torque predictions remain well-calibrated, we define two mean-squared error objectives:
\(\mathcal{L}_{\rm action}(\theta)
\mathbb{E}_{\boldsymbol{Z}_t,\boldsymbol{\epsilon},\alpha}\Bigl\|
\boldsymbol{v}_\theta(\boldsymbol{Z}_t^\alpha, o_t)_{A} - (\boldsymbol{\epsilon}_{A} - \boldsymbol{A}_t)
\Bigr\|_2^2\),
\(\mathcal{L}_{\rm torque}(\theta)
\mathbb{E}_{\boldsymbol{Z}_t,\boldsymbol{\epsilon},\alpha}\Bigl\|
\boldsymbol{v}_\theta(\boldsymbol{Z}_t^\alpha, o_t)_{T} - (\boldsymbol{\epsilon}_{T} - \boldsymbol{T}_t)
\Bigr\|_2^2\),
where $\boldsymbol{v}_\theta(\cdot)_A$ and $\boldsymbol{v}_\theta(\cdot)_T$ denote, respectively, the action- and torque-components of the model’s output, and $\boldsymbol{\epsilon}_A$, $\boldsymbol{\epsilon}_T$ are the corresponding slices of the noise. However, unlike the commonly adopted method that multiple types of outputs are predicted through separate modules or shared weights with different projection heads, to save cost and leverage pre-trained weights, we use a single linear layer instead that outputs concatenated action and torque predictions together, then split them back for their respective losses. 
We adopt the combination of the two losses:
\(\mathcal{L}_{\rm joint}(\theta)
= \mathcal{L}_{\rm action}(\theta)
+ \beta\,\mathcal{L}_{\rm torque}(\theta)\),
where \(\beta\) is a weighting factor  balancing action fidelity and torque accuracy.

\begin{figure}
    \centering
    \includegraphics[width=0.9\linewidth]{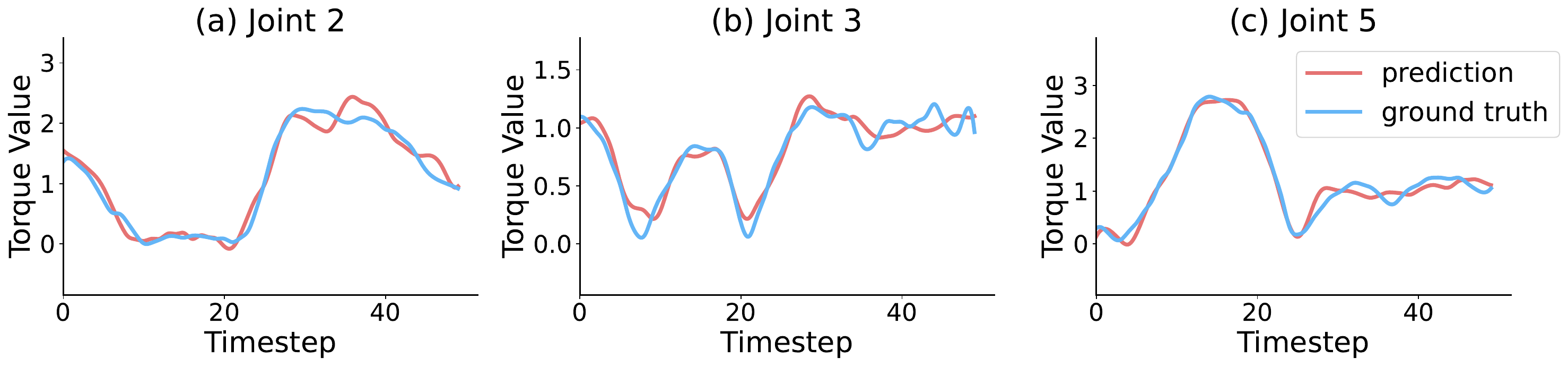}
    \vspace{-8pt}
    \caption{Future torque signal prediction for joints.}
    \label{fig:eff_pred}
    \vspace{-0.7cm}
\end{figure}

\textbf{Empirical Results.}
To evaluate the precision of future torques across joints predicted by the action-torque diffusion method, we compare with ground-truth values in validation data. As shown in Figure~\ref{fig:eff_pred}, the predicted torques highly comply with the ground-truth variations, which implies that through the proposed joint diffusion method, the model is able to successfully sense the future changes. This ability will further enable the model to yield better actions (results are shown in Sec.~\ref{sec: Quantitative Results}), because joint torque-action prediction strategy strengthens the model’s understanding of contact dynamics by learning the causal relationship between actions and the resulting torque responses. By learning a unified action–torque representation, the model aligns proprioceptive signals with intended motor commands, which enhances performance in contact-rich scenarios.

\vspace{-9pt}
\section{Experiment}
\label{sec:exp}
\vspace{-6pt}
\subsection{Experimental Setup}
\vspace{-5pt}
\textbf{Hardware Platform.} We use the Cobot Magic ALOHA, a dual-arm robot with 7 degrees of freedom per arm. It is equipped with three D435 depth cameras: two on the wrists and one front-facing. The joint torques are derived from the robot’s motors based on the electrical currents supplied to them. Each motor has a specific \textit{current-to-torque constant} \(k_t\), which relates the current \(i\) to the generated torque $\tau$, calculated as \(\tau = k_t i\). By measuring the current supplied to each motor, we can accurately estimate the joint torque in real-time without the need for external force sensors.

\textbf{Baselines.} We evaluate against strong baselines in robotic manipulation: ACT~\citep{ACT}, RDT~\citep{RDT}, and $\pi_0$~\citep{pi0}. All models are fine-tuned from publicly available pretrained weights on our collected dataset under the same experimental setup. ACT leverages action chunking with transformers, while RDT and $\pi_0$ are two state-of-the-art VLA models with strong cross-task performance.


\vspace{-5pt}
\subsection{Quantitative Results}
\vspace{-5pt}
\label{sec: Quantitative Results}
We evaluate the baseline model together with multiple ways to incorporate torque signals to $\pi_0$. Specifically, we adopt (1) \textit{DePost-1 Token Architecture} in Sec.~\ref{sec:obs} to embed immediate and past torque observations, denoted as \textbf{$\pi_0$+obs}; (2) the \textit{unified training objective} in Sec.\ref{sec:obj}, denoted as \textbf{$\pi_0$+obj}; and (3) their combination, denoted as \textbf{$\pi_0$+obs+obj}. We conduct real-world experiments across 10 tasks—5 contact-rich and 5 regular. Results in Table~\ref{tab:Quantitative Results} show that both torque observations and torque-based objectives benefit VLA models. The combined approach leverages the strengths of both, yielding the best overall performance.
Also, torque signals improve not only contact-rich tasks but also tasks where torque appears less relevant, indicating their utility across diverse scenarios.
\begin{table*}[ht]
  \centering
  \vspace{-0.4cm}
  \renewcommand{\arraystretch}{0.9}
  \setlength{\tabcolsep}{3pt}
  \resizebox{1\linewidth}{!}{
  \begin{tabular}{l|c|c|c|c|c}
    \toprule
    Task & \multicolumn{5}{c}{Contact-rich Task}\\
    \midrule
    Method & Button Pushing & Charger Plugging & USB Plugging & Socket Unplugging & Door Handle Turning\\
    \midrule
    ACT & 2/20 & 0/20 & 0/20 & 12/20 & 0/20 \\
    RDT & 4/20 & 1/20 & 0/20 & 10/20 & 0/20 \\
    $\pi_0$ & 5/20 & 0/20 & 0/20 & 16/20 & 2/20 \\ 
    \rowcolor{gray!15}$\pi_0$ + obs & 15/20 & 16/20 & 15/20 & \textbf{19/20} & 13/20 \\
    \rowcolor{gray!20}$\pi_0$ + obj & 11/20 & 10/20 & 12/20 & \textbf{19/20} & 12/20 \\
    \rowcolor{gray!25} $\pi_0$ + obs + obj & \textbf{18/20} & \textbf{17/20} & \textbf{17/20} & \textbf{19/20} & \textbf{15/20} \\
    \bottomrule
    \toprule
    Task & \multicolumn{5}{c}{Regular Task}\\
    \toprule
    Method & Bottle Pick and Place & Liquid Pouring & Stacking Cubes & Push-to-Position & Opening a Drawer\\
    \midrule
    ACT & 15/20 & 13/20 & 12/20 & 13/20& 16/20\\
    RDT & 17/20 & \textbf{17/20} & 12/20 & 15/20& 16/20\\
    $\pi_0$ & 17/20 & 16/20 & 17/20 & 16/20 & \textbf{19/20}\\ 
    \rowcolor{gray!15}$\pi_0$ + obs & 18/20 & 16/20 & \textbf{18/20} & 16/20 & \textbf{19/20}\\
    \rowcolor{gray!20}$\pi_0$ + obj & 17/20 & 16/20 & 17/20 & 16/20 & 18/20\\
    \rowcolor{gray!25} $\pi_0$ + obs + obj & \textbf{19/20} & \textbf{17/20 }& 17/20 & \textbf{18/20} & 18/20\\
    \bottomrule
  \end{tabular}
  }
  \caption{\textbf{Quantitative Results.} Success rates across 5 contact-rich tasks and 5 regular tasks, each evaluated 20 trials. Our method consistently outperforms baselines, especially in contact-rich tasks.}
  \label{tab:Quantitative Results}
  \vspace{-0.5cm}
\end{table*}

\vspace{-5pt}
\subsection{Visualization}
\begin{figure}
    \centering
    \includegraphics[width=\linewidth]{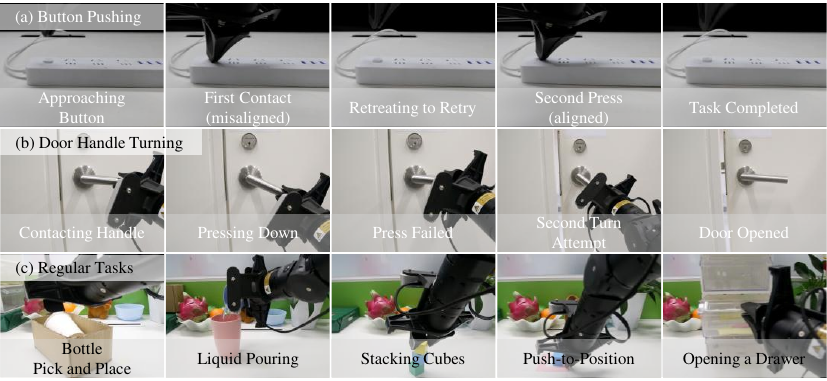}
    \vspace{-15pt}
    \caption{\textbf{Visualization.} (a) Button Pushing: First attempt fails due to misalignment; second succeeds. (b) Door Handle Turning: Initial turn fails; second opens the door. (c) 5 regular tasks.}
    \label{fig:visualization}
    \vspace{-0.5cm}
\end{figure}

\vspace{-5pt}
We visualize part of contact-rich and regular tasks the proposed method can achieve. Regarding contact-rich tasks, as shown in Figure~\ref{fig:visualization}. When detecting a failed attempt due to abnormal changes in joint torque caused by misalignment or slippage (the second and third images in Figure~\ref{fig:visualization}(a)-(b)), leveraging torque feedback, the robot autonomously retries the motion and successfully completes the task on the second attempt (the fourth and fifth images in Figure~\ref{fig:visualization}(a)-(b)). Additionally, with torque signals, the robot can complete various regular tasks with high precision (Figure~\ref{fig:visualization}(c)).

\vspace{-5pt}
\subsection{Cross Model}
\vspace{-5pt}
\begin{wrapfigure}{r}{0.5\textwidth}
\vspace{-35pt}
\centering
  \resizebox{1\linewidth}{!}{
    \renewcommand{\arraystretch}{0.8}
  \begin{tabular}{l|c|c|c}
    \toprule
    Method & Button Pushing & Charger Plugging & Bottle Pick and Place\\
    \midrule
    RDT & 4/20 & 1/20 & 17/20 \\
    \rowcolor{gray!25} RDT + obs + obj & \textbf{16/20} & \textbf{15/20} & \textbf{19/20}\\
 \bottomrule
  \end{tabular}   
  }
  \captionof{table}{\textbf{Cross Model Results.} Success rates across contact-rich and regular tasks on RDT.}
  \label{tab:rdt}
  \vspace{-0.6cm}
\end{wrapfigure}
To evaluate the generalization capability of torque observations and torque-based objectives across different VLA models, we conduct experiments on RDT~\citep{RDT} across both contact-rich and regular tasks. As shown in Table~\ref{tab:rdt}, incorporating both torque observations and the torque-based objective leads to significantly improved performance. These results suggest that the torque integration strategies introduced in Sec.\ref{sec:obs} and Sec.\ref{sec:obj} generalize well to other VLA models.

\vspace{-10pt}
\subsection{Cross Embodiment}
\vspace{-5pt}
\begin{wrapfigure}{l}{0.45\textwidth}
\centering
\vspace{-0.7cm}
\includegraphics[scale=0.6]{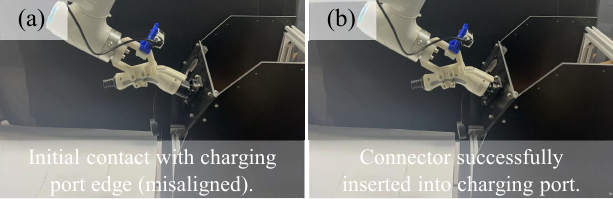}
\caption{Cross Embodiment Visualization.}
\label{fig:cross_embodiment}
\vspace{-0.4cm}
\end{wrapfigure}

To evaluate the generalization capability of our method across different robotic embodiments, we conduct experiments on a ROKAE SR robotic arm. As shown in Figure~\ref{fig:cross_embodiment}, the robot performs an electric vehicle charger insertion task. 
After detecting a failed insertion attempt using torque feedback (Figure~\ref{fig:cross_embodiment}(a)), the robot successfully completes the task on its second attempt (Figure~\ref{fig:cross_embodiment}(b)).

\vspace{-9pt}
\section{Conclusion}
\vspace{-6pt}
In this paper, we analyze joint torques as effective indicators of end-effector status and explore how to best incorporate them into VLA models. We find that encoding immediate and historical torques as a single decoder token yields the best results. Further, jointly predicting actions and torques with a unified diffusion loss improves performance. Experiments on contact-rich and regular tasks confirm the effectiveness and generalization of both torque-based enhancements.

\section{Limitations}
\vspace{-6pt}
\label{limitations}
The method relies on accurate torque estimation from internal motor currents. This estimation can be affected by motor calibration, sensor noise, or thermal drift, potentially degrading performance in prolonged or high-load tasks. Moreover, while torque signals prove valuable, it remains unclear how scalable the framework is when extending to other physical modalities like tactile sensing or temperature, especially under shared token budgets in transformer architectures. Future work is needed to evaluate robustness in more diverse, real-world scenarios, as well as to explore the alignment and integration of richer multimodal signals.



\bibliography{main}  

\clearpage
\newpage

\setcounter{page}{1}
\begin{center}
    {\LARGE \textbf{Elucidating the Design Space of Torque-aware}}\\
    {\LARGE \textbf{Vision-Language-Action Models}}
\end{center}

\begin{appendices}
\section{Appendix}
In this appendix, we provide a comprehensive elaboration of the technical, experimental, and implementation details of our study. Sec.~\ref{sec:app-Additional Visualizations} presents additional qualitative visualizations to supplement the main text, highlighting the system's performance in various scenarios. Sec.~\ref{sec:app-Detailed Wrench‑to‑Torque Mapping for a 7‑DOF Manipulator} meticulously derives the wrench-to-torque mapping for a 7-DOF manipulator, including the full spatial Jacobian, joint-specific partitions, and a quasi-static simplification for efficient torque computation. Sec.~\ref{sec:app-secA.3} and~\ref{sec:app-secA.4} detail the experimental protocols for the torque-integration and torque-history encoding strategies, ensuring reproducibility. Sec.~\ref{sec:app-secA.5} outlines the implementation specifics of the joint action-torque diffusion objective, clarifying how torque prediction is achieved alongside action generation. Sec.~\ref{sec:app-secA.6} provides further insights into the experimental setup, quantitative results, and cross-model as well as cross-embodiment evaluations, ensuring a robust understanding of our results. Sec.~\ref{sec:app-secA.7} describes the architectural specifications of the baseline VLA models ($\pi_0$ and RDT). Sec.~\ref{sec:app-system efficiency} evaluates the system’s efficiency in terms of training and inference, demonstrating that our torque-aware enhancements do not significantly impact computational performance. Sec.~\ref{Sec:beta} ablates the loss weight $\beta$ for the joint action–torque diffusion objective on Button Pushing.
Sec.~\ref{Sec:torque aggregation} compares torque-history aggregation (MLP, RNN, attention) under identical settings.
Finally, Sec.~\ref{Sec:torque interpretation preprocessing} clarifies torque-signal usage and preprocessing.

\subsection{Additional Visualizations}
\label{sec:app-Additional Visualizations}
\begin{figure}[h]
    \centering
    \includegraphics[width=\linewidth]{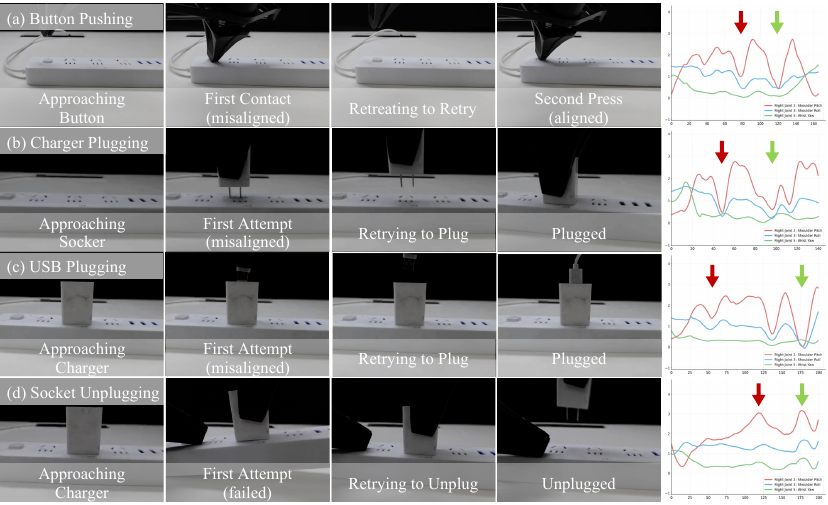}

    \caption{\textbf{Visualization of contact-rich tasks.} The last column visualizes the torque response of some joints during the task. For each task, the first failed attempt is marked with a \textcolor[HTML]{b80000}{red} arrow, and the final successful attempt is marked with a \textcolor[HTML]{87ca46}{green} arrow.}
    \label{fig:app_effort_task}

\end{figure}

\begin{figure}
    \centering
    \includegraphics[width=\linewidth]{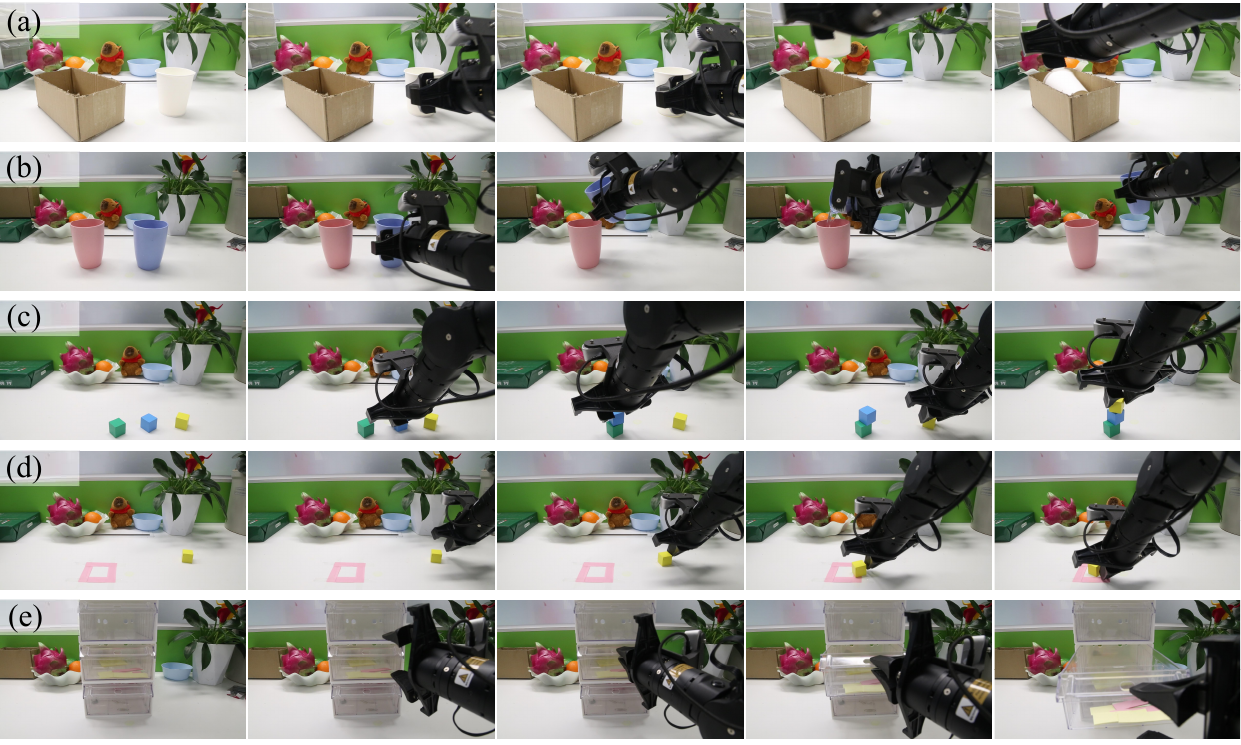}

    \caption{\textbf{Visualization of general tasks.} (a) Bottle pick and place. (b) Liquid pouring. (c) Stacking cubes. (d) Push-to-position. (e) Opening a drawer.}
    \label{fig:app_general_task}

\end{figure}

We provide additional visualizations of the model's execution process for the tasks described in the main text in Figures~\ref{fig:app_effort_task} and~\ref{fig:app_general_task}. For the contact-rich tasks, the torque response of the Shoulder Pitch, Shoulder Roll, and Wrist Yaw joints during execution is plotted in the last column, similar to Figure~\ref{teaser}. We have marked the torque changes corresponding to failed and successful attempts for each task. See our \href{https://zzongzheng0918.github.io/Torque-Aware-VLA.github.io/}{project page} for complete videos.

\subsection{Detailed Wrench‑to‑Torque Mapping for a 7‑DOF Manipulator}
\label{sec:app-Detailed Wrench‑to‑Torque Mapping for a 7‑DOF Manipulator}
Below we expand the derivation in Sec.\ref{sec:method}, make explicit the structure of the Jacobian, and specialize it to a 7‑DOF arm in which the first six joints are revolute actuators (shoulder→wrist) and the 7‑th joint is the gripper’s open/close degree of freedom.

\subsubsection{Full Spatial Jacobian}

For a serial arm with $n$ joints, the geometric Jacobian $J(\boldsymbol{q}) \in \mathbb{R}^{6 \times n}$ stacks the linear and angular velocity components:
\begin{equation}
J(\boldsymbol{q}) =
\begin{bmatrix}
J_v(\boldsymbol{q}) \\
J_\omega(\boldsymbol{q})
\end{bmatrix}, \quad
J_v, J_\omega \in \mathbb{R}^{3 \times n}.
\end{equation}

For a revolute joint $j$ with axis $\hat{\boldsymbol{z}}_j$ and origin $\boldsymbol{p}_j$ (in the base frame):
\begin{equation}
\begin{aligned}
J_{v[:,j]} &= \hat{\boldsymbol{z}}_j \times (\boldsymbol{p}_e - \boldsymbol{p}_j), \\
J_{\omega[:,j]}& = \hat{\boldsymbol{z}}_j,
\end{aligned}
\end{equation}
where $\boldsymbol{p}_e$ is the end-effector position.

\subsubsection{Partition for the 7-DOF Arm} 

Let the joint order be:
\begin{equation}
[q_1, \ldots, q_6, q_7] = 
\underbrace{[\text{Sh. Yaw, Sh. Pitch, Sh. Roll, Elb., Wr. Yaw, Wr. Pitch},}_{\text{arm (1-6)}} \quad 
\underbrace{\text{Gripper open/close}]}_{7}.
\end{equation}

Then:
\begin{equation}
J(\boldsymbol{q}) = [J_{\text{arm}}(\boldsymbol{q}) ; J_{\text{grip}}(\boldsymbol{q})], \quad
J_{\text{arm}} \in \mathbb{R}^{6 \times 6}, \quad J_{\text{grip}} \in \mathbb{R}^{6 \times 1}.
\end{equation}

\textbf{Gripper column.} The gripper's opening motion does not change the Cartesian pose of the tool‑centre‑point (TCP), so:
\begin{equation}
J_{\text{grip}}(\boldsymbol{q}) = \boldsymbol{0}_{6 \times 1} \Rightarrow \tau_{\text{ext},7} = 0.
\end{equation}
Consequently, forces at the TCP do not back‑propagate torque to $q_7$ :

\begin{equation}
\tau_{\text{ext},7} = J_{\text{grip}}^\top \boldsymbol{F}_{\text{ext}} = 0.
\end{equation}
 
\textbf{Arm columns 1:6.} Each column is computed using the revolute formula above. For clarity we show the symbolic structure:

\begin{equation}
J_{\text{arm}} =
\begin{bmatrix}
\hat{z}_1 \times (\boldsymbol{p}_e - \boldsymbol{p}_1) & \dots & \hat{z}_6 \times (\boldsymbol{p}_e - \boldsymbol{p}_6) \\
\hat{z}_1 & \dots & \hat{z}_6
\end{bmatrix}.
\end{equation}

\subsubsection{Wrench--Torque Projection}
Given the external wrench $\boldsymbol{F}_{\text{ext}} = [\boldsymbol{f}, \boldsymbol{m}]^\top \in \mathbb{R}^6$:
\begin{equation}
\boldsymbol{\tau}_{\text{ext}} = J^\top(\boldsymbol{q}) \boldsymbol{F}_{\text{ext}} = \begin{bmatrix} J_{\text{arm}}^\top\boldsymbol{F}_{\text{ext}} \\ 0 \end{bmatrix} \in \mathbb{R}^7,
\end{equation}
where
\(
\tau_{\text{ext},1:6} = J_{\text{arm}}^\top \boldsymbol{F}_{\text{ext}},
\)
 and
\(
\tau_{\text{ext},7} = 0.
\)

\subsubsection{Quasi-static Simplification}
Under low-velocity manipulation $\dot{\boldsymbol{q}}, \ddot{\boldsymbol{q}} \approx 0$, Eq.~\eqref{eq:dynamics} simplifies to:
\begin{equation}
\boldsymbol{\tau}_{\text{measured}} \approx \boldsymbol{G}(\boldsymbol{q}) + \begin{bmatrix} J_{\text{arm}}^\top \boldsymbol{F}_{\text{ext}} \\ 0 \end{bmatrix}.
\end{equation}
Subtracting $\boldsymbol{G}(\boldsymbol{q})$ yields the external component $\delta \boldsymbol{\tau}_{\text{ext}}$, where only joints 1--6 are informative for contact detection. Large residuals in those joints directly indicate contact onset, direction, and magnitude, while small residuals in 
$q_7$ confirm that gripper actuation alone does not contribute contact‑induced torques.

\subsubsection{Practical Notes for Implementation}
\begin{itemize}
\item \textbf{Gravity term $\boldsymbol{G}$:} Estimate via CAD model; recalibrate with payload when grasping heavy objects.
\item \textbf{Torque from motor currents:} Use manufacturer-provided $k_t$ with thermal compensation.
\item \textbf{Contact detection:} Threshold $\delta \tau$ values per joint to detect abnormal force events.
\end{itemize}

This detailed formulation clarifies the exact Jacobian structure, shows why the gripper joint is torque-insensitive to TCP forces, and provides concrete implementation guidance suitable for reproducibility in a top-tier robotics venue.

\subsection{Experimental Protocols for Sec.~\ref{subsec:obs_where}: Torque‑Integration Architectures}
\label{sec:app-secA.3}

\textbf{Experiment about Comparison between Architectures.}
\label{app:mlp_arch}
For the \textbf{Enc} and \textbf{DePost} architectures (Figure~\ref{fig:where_embed}(a)(c)), we randomly initialize a MLP to project the effort token into the latent space. This MLP is structured with layers mapping from an input dimension of 14 (effort dim) to \(2 \times width\), followed by a Swish activation, and then mapping from \(2 \times width\) to \(width\). The \(width\) here corresponds to the model's internal dimension: 2048 for the conditioning encoder in the \textbf{Enc} architecture and 1024 for the diffusion decoder in the \textbf{DePost} architecture.
The state input of \(\pi_0\) is composed of the 14-dimensional joint positions, followed by 18 dimensions of zero-padding. For the \textbf{DePre} architecture (Figure~\ref{fig:where_embed}(b)), we place the 14-dimensional joint efforts into the last 14 positions of this 32-dimensional state.

\textbf{Experiment about Better Input Alignment.} 
HSIC is a powerful nonparametric measure capable of detecting complex nonlinear relationships between variables without requiring assumptions about their distribution. Normalized HSIC provides a value between 0 and 1, where higher values indicate stronger statistical dependence.
To evaluate modality alignment, we analyze the \(\pi_0\) model trained using the \textbf{DePost} method on the Button Pushing task. We process input tokens obtained from frames randomly sampled from the training dataset of this task through the model's transformer backbone (18 layers total). We then extract the intermediate representations, specifically the hidden states at the output of the 12th layer, corresponding to these input tokens. For the HSIC computation, the number of these extracted intermediate token representations used for each modality (action, angle, torque, image, and text instruction) is downsampled to 32. The normalized HSIC is then computed pairwise between each combination of modalities using RBF kernels, shown in Figure~\ref{fig:hsic}.

\textbf{Experiment about Sensitivity of Decoder.}
To assess the sensitivity of the encoder and decoder to input variations, we add additive Gaussian noise with a standard deviation of 0.1 to the input tokens. For the encoder, noise is applied to all input tokens, whereas for the decoder, noise is applied specifically to the state token (the first token in the input sequence). The results are presented in Table~\ref{tab:noised}.

\subsection{Experimental Protocols for Sec.~\ref{subsec:obs_his}: Torque‑History Encoding}
\label{sec:app-secA.4}

\textbf{Experiment about Comparison between Architectures.}
For historical effort input, we uniformly sampled 10 frames from the past 2 seconds, including the current frame.
In the H Tokens configuration, each of these 14-dimensional effort tokens is processed independently using an MLP with the same architecture as described in Appendix \ref{app:mlp_arch}, which projects it into the latent space.
Conversely, in the 1 Token configuration, the historical effort from all 10 frames is flattened and concatenated into a single 140-dimensional token before being fed into the MLP.

\textbf{Experiment about Better Input Pattern Completeness.}
To investigate the completeness of the input pattern, we introduced an additional token into the input token sequence. This token was sampled from a standard normal distribution and had the same shape as other tokens in the sequence. We then modified the input and autoregressive masks accordingly, following the masking patterns applied to the effort tokens to integrate this new token into the sequence processing. Table~\ref{tab:disrupted} shows the results of the experiment.

\subsection{Implementation of the Joint Action‑Torque Diffusion Objective (Sec.~\ref{sec:obj})}
\label{sec:app-secA.5}
To enable the model to simultaneously output future effort (torque) for supervision alongside action predictions, we expanded the dimension of the action input and output projection linear layer of the original model (Figure~\ref{fig:future curve}). When loading the pre-trained weights, we initialized the portion of the modified weight matrix corresponding to the original action output dimensions with the pre-trained values. The remaining weights, which correspond to the newly added dimensions for future effort prediction, were initialized using smaller values. This initialization strategy is designed to minimally affect the original pre-trained behavior initially, allowing the model to gradually learn the new prediction task during the finetuning process. The predicted future effort sequence has a length \(H=50\) steps, matching the action chunk length. Figure~\ref{fig:eff_pred} shows an example inference from the model trained on the Button Pushing task. The figure plots the model's predicted future effort per frame against the corresponding ground truth for three selected axes, using an observation frame sampled from the task's validation data as input.

\subsection{Additional Details in Sec.~\ref{sec:exp}}
\label{sec:app-secA.6}

\subsubsection{Details in Experimental Setup}
In the \(\pi_0\) experiments, we followed the original setup using images from three viewpoints as input: top, left wrist, and right wrist. All \(\pi_0\) experiments were based on its publicly available pre-trained checkpoint and finetuned both encoder and decoder using LoRA. Training was performed for 30k gradient steps on 4 × NVIDIA L20 GPUs. The RDT experiments used the same GPU setup for full parameter training for 40k gradient steps. For ACT, we use 600 training epochs with a chunk size of 32. All baseline models use AdamW optimizer. Inference was performed using an onboard RTX 4090 GPU. All variants of \(\pi_0\) used an inference action horizon of 50 steps, and RDT used 64 steps. Other settings remained consistent with the original implementation. For all tasks, 400 demonstrations were collected using teleoperation.

\subsubsection{Details about Quantitative Results}
In the experiments (Table~\ref{tab:Quantitative Results}), for $+obj$ settings, we set the value of \(\beta\) to 1. For $+obs+obj$ settings, we set the value of $\beta$ to 0.1. For the ACT and RDT models, their inference action horizons were 8 and 64 steps, respectively.

\subsubsection{Details about Cross Model Results}
\begin{wrapfigure}{r}{0.4\textwidth}
\vspace{-1.5cm}
    \centering
    \includegraphics[scale=1]{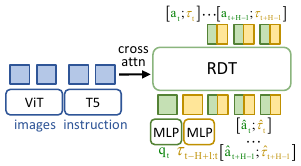}
    \caption{Architecture of RDT+obs+obj model.}
    \label{fig:app_cross_model}
    \vspace{-0.4cm}
\end{wrapfigure}

Regarding the RDT+obs+obj model (Figure~\ref{fig:app_cross_model}, Table~\ref{tab:rdt}), following the model's approach for language and image adapters, we implement the effort projector as a two-layer MLP with a single GELU activation. This MLP maps the effort input into a 2048-dimensional vector, matching the width of RDT's transformer backbone. The projected effort token is then concatenated after the state token. This combined sequence is further concatenated with the noisy action token, forming the input for the denoising process in RDT. We extended the state space input and output projectors and loaded pre-trained weights in a manner similar to that used in \(\pi_0\).

\subsubsection{Details about Cross Embodiment Results}

\begin{figure}
    \centering
    \includegraphics[width=\linewidth]{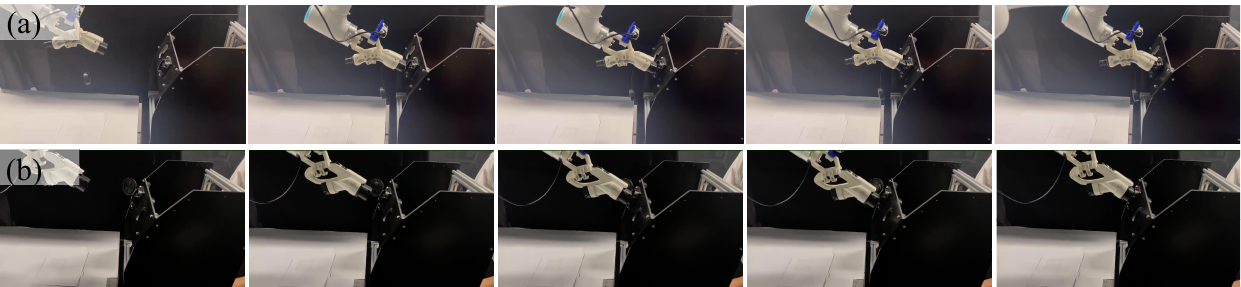}

    \caption{\textbf{Cross-Embodiment Task Execution: Charging Connector Insertion.} (a) The robotic manipulator successfully inserts a fast-charging connector into the charging port. 
(b) The robotic manipulator inserts a slow-charging connector into the charging port. }
    \label{fig:app_cross_embodiment}

\end{figure}

To further assess the generalization capability of our torque-aware VLA model across different robotic embodiments, we conducted cross-embodiment experiments using the ROKAE SR robotic arm. Specifically, we trained the $\pi_0$+obs+obj with 200 demonstrations of the robot inserting a fast-charging connector into the charging port, using torque feedback to guide the insertion process. The model was trained for 50K gradient steps, and as shown in Figure~\ref{fig:app_cross_embodiment}(a), it successfully achieved the fast-charging insertion task with high reliability. To further evaluate generalization, we altered the task setup by replacing the fast-charging port with a slow-charging port. Without any architecture modification, the model was able to adapt seamlessly, successfully inserting the slow-charging connector into the port, as demonstrated in Figure~\ref{fig:app_cross_embodiment}(b). This result highlights the model's capacity to generalize torque-aware manipulation strategies to new end-effector configurations, demonstrating robust cross-embodiment performance.

\subsection{Architectural Specifications of Baseline VLA Models ($\pi_0$ and RDT)}
\label{sec:app-secA.7}
\(\pi_0\) is a VLA model built by adding a 300M action expert on top of the PaliGemma 3B pretrained VLM backbone and trained on a large-scale cross-embodiment robot dataset. The action expert and the VLM are two independent sets of weights within a single transformer, interacting only through the self-attention layers. During each inference step, the model receives three RGB images, a language instruction, and the robot's proprioceptive state. The images and language instruction are fed directly into the VLM as conditions. Proprioception is concatenated with a noisy action chunk of length 50 and input to the action expert. \(\pi_0\) uses Conditional Flow Matching to model the continuous action distribution. The action expert outputs a vector field, and the final action chunk is generated by performing 10 integration steps on this vector field.

RDT (Robotics Diffusion Transformer) is a 1B-parameter Diffusion Transformers (DiTs) model pre-trained using 1M multi-robot trajectories (including 46 datasets). It utilizes a physically interpretable 128-dimensional unified observation and action space, encoding low-dimensional inputs using MLPs with Fourier features. During each inference step, the model's inputs include proprioception, a noisy action chunk (size 64), and the diffusion timestep. Language instruction and observations (including 2 history frames of three images, proprioceptive state, and control frequency) are input to the model as conditions. Images are processed by SigLIP, while language is processed by T5-XXL, this conditional information is alternately injected into different layers of the DiTs using cross attention. The model predicts the denoised action chunk, and the final action is produced through 5 denoising steps.

%

\subsection{System Efficiency}
\label{sec:app-system efficiency}
We use the Button Pushing task as an example to compare the training and inference times for \(\pi_0\), \(\pi_0\)+obs, \(\pi_0\)+obj, and \(\pi_0\)+obs+obj. The training speed is reported as the average throughout the training process measured on 4 × NVIDIA L20 GPUs, and the inference speed is averaged over 10 runs on an RTX 4090 GPU. The results are shown in Table \ref{tab:training_time} and Table \ref{tab:inference_time}, which indicate that the proposed designs do not significantly affect efficiency.


\begin{table}[ht]
\centering
\begin{minipage}[t]{0.49\textwidth}
  \centering
  \setlength{\tabcolsep}{2pt}
  \fontsize{7pt}{7.5pt}\selectfont
  \renewcommand{\arraystretch}{0.7}
  \begin{tabular}{l|c}
    \toprule
    Model Variant & Training Time (s/iter) \\
    \midrule
    \(\pi_0\) & 1.703 \\
    \(\pi_0+\text{obs}\) & 1.628 \\
    \(\pi_0+\text{obj}\) & 1.648 \\
    \(\pi_0+\text{obs}+\text{obj}\) & 1.640 \\
    \bottomrule
  \end{tabular}
  \vspace{0.1cm}
  \caption{Training Time for Different Designs}
  \label{tab:training_time}
\end{minipage}
\hfill
\begin{minipage}[t]{0.49\textwidth}
  \centering
  \setlength{\tabcolsep}{2pt}
  \fontsize{7pt}{7.5pt}\selectfont
  \renewcommand{\arraystretch}{0.7}
  \begin{tabular}{l|c}
    \toprule
    Model Variant & Inference Time (ms) \\
    \midrule
    \(\pi_0\) & 90.70 \\
    \(\pi_0+\text{obs}\) & 90.81 \\
    \(\pi_0+\text{obj}\) & 90.61 \\
    \(\pi_0+\text{obs}+\text{obj}\) & 93.96 \\
    \bottomrule
  \end{tabular}
  \vspace{0.1cm}
  \caption{Inference Time for Different Designs}
  \label{tab:inference_time}
\end{minipage}
\end{table}

\subsection{Ablation Studies for Hyperparameter $\beta$}
\label{Sec:beta}
We test  $\beta$ on $\pi_0$+obj and $\pi_0$+obs+obj in \textit{button pushing} task (Tab.~\ref{beta}).
For $\pi_0$+obj: Success rises from 6/20 at $\beta=0.01$ and reaches a plateau from 0.2 to 1, so we let $\beta=1$.
For $\pi_0$+obs+obj: Peak success (18/20) occurs at lower $\beta$ before dropping at higher values, so we set it as 0.1 to balance the two newly introduced components: auxiliary loss and torque observation modality.

\begin{table}[ht]
\centering
\begin{minipage}[t]{0.49\textwidth}
  \centering
  \setlength{\tabcolsep}{2pt}
  \fontsize{7pt}{7.5pt}\selectfont
  \renewcommand{\arraystretch}{0.7}
  \begin{tabular}{l|ccccc}
    \toprule
    $\beta$ & 0.01 & 0.1 & 0.2 & 0.5 & 1 \\
    \midrule
    \(\pi_0+\text{obj}\) & 6/20 & 8/20 & 10/20 & 9/20 & \textbf{11}/20\\
    \(\pi_0+\text{obs}+\text{obj}\) & 14/20 & \textbf{18}/20 & \textbf{18}/20 & 15/20 & 12/20 \\
    \bottomrule
  \end{tabular}
  \vspace{0.1cm}
  \caption{Ablation study on $\beta$.}
  \label{beta}
\end{minipage}
\hfill
\begin{minipage}[t]{0.49\textwidth}
  \centering
  \setlength{\tabcolsep}{2pt}
  \fontsize{7pt}{7.5pt}\selectfont
  \renewcommand{\arraystretch}{0.7}
  \begin{tabular}{l|ccc}
    \toprule
    Aggregation & MLP& RNN &Attention \\
    \midrule
    \(\pi_0+\text{obs}\) & \textbf{15}/20 & 7/20 & 13/20  \\
    \(\pi_0+\text{obs}+\text{obj}\) & \textbf{18}/20 & 10/20 & 17/20  \\
    \bottomrule
  \end{tabular}
  \vspace{0.1cm}
  \caption{Inference Time for Different Designs}
\label{agg}
\end{minipage}
\end{table}

\subsection{Ablation Studies for Torque Aggregation Methods}
\label{Sec:torque aggregation}

We compare three aggregation methods for torque history in \textit{button pushing} task, using the same hyperparameters. As shown in Table~\ref{agg}, a simple MLP outperforms others. We attribute it to the parameter efficiency: with limited fine-tuning data, more complex sequence models (RNN/attention) tend to underfit.

\subsection{Torque Signal Interpretation and Preprocessing}
\label{Sec:torque interpretation preprocessing}
\begin{figure}
    \centering
    \includegraphics[width=0.5\linewidth]{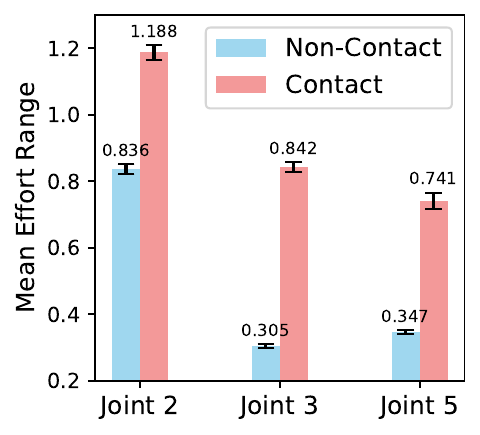}
    \caption{Torque Signals}
    \label{effort}
\end{figure}

We use torque signals from \textbf{all 7 joints}, not only the end-effector. Due to robot dynamics, some joints—especially near the shoulder (e.g., Joint 2 in Figure~\ref{teaser}(b)) exhibit variation during non-contact motion (e.g., 20–36 timestamps in Figure~\ref{teaser}(a)). These changes are not noise, and are much smaller than contact-induced signals (Figure~\ref{effort}). Additionally, to ensure stability and keep meaningful patterns, we normalize each joint using training data statistics, as is done for states and actions.

\end{appendices}

\end{document}